\documentclass[lettersize, journal]{IEEEtran}
\usepackage{amsmath,amsfonts}
\usepackage{array}
\usepackage[caption=false,font=normalsize,labelfont=sf,textfont=sf]{subfig}
\usepackage{textcomp}
\usepackage{stfloats}
\usepackage{xurl}
\usepackage{verbatim}
\usepackage{graphicx}
\usepackage{cite}
\usepackage{soul}
\usepackage[dvipsnames]{xcolor}
\usepackage{color}
\usepackage{microtype}
\usepackage{booktabs}
\usepackage{lmodern}
\usepackage{amssymb}
\usepackage{amsmath}
\usepackage{algpseudocode}
\usepackage{enumitem}
\usepackage{algorithm}
\usepackage{booktabs}
\usepackage{tcolorbox}
\usepackage{framed}
\usepackage[flushleft]{threeparttable}
\usepackage[normalem]{ulem}
\useunder{\uline}{\ul}{}
\usepackage{makecell}
\usepackage{tablefootnote}
\usepackage{multirow}
\usepackage{tabulary}
\begin{document}

\title{Beyond RMSE and MAE: Introducing EAUC to unmask hidden bias and unfairness in dyadic regression models}

\author{
\IEEEauthorblockN{Jorge Paz-Ruza, Amparo Alonso-Betanzos, Bertha Guijarro-Berdiñas, Brais Cancela, and Carlos Eiras-Franco}

\thanks{Revised manuscript received \today 

All authors are with the LIDIA Group, CITIC, Universidade da Coruña, 15701 Spain. E-mails: \{j.ruza, amparo.alonso.betanzos, berta.guijarro, brais.cancela, carlos.eiras.franco\}@udc.es

This research work has been funded by MICIU/AEI/10.13039/501100011033 and ESF+ (FPU21/05783), \textit{ERDF A way of making Europe} (PID2019-109238GB-C22, PID2023-147404OB-I00), ERDF EU (PID2021-128045OA-I00), and by the Xunta de Galicia (ED431C 2022/44) with the European Union ERDF funds. CITIC, as Research Center accredited by Galician University System, is funded by ``Consellería de Cultura, Educación e Universidade from Xunta de Galicia'', supported in an 80\% through ERDF Operational Programme Galicia 2021-2027, and the remaining 20\% by ``Secretaría Xeral de Universidades'' (ED431G 2023/01).
}
}%



\maketitle

\colorlet{ChangesColor}{SkyBlue!0}

\sethlcolor{ChangesColor}

\newcommand{\mathhl}[1]{\colorbox{ChangesColor}{$\displaystyle #1$}}

\IEEEpeerreviewmaketitle

\begin{abstract}
Dyadic regression models, which output real-valued predictions for pairs of entities, are fundamental in many domains (e.g. obtaining user-product ratings in Recommender Systems) and promising and under exploration in others (e.g. tuning patient-drug dosages in precision pharmacology). In this work, we prove that non-uniform observed value distributions of individual entities lead to severe biases in state-of-the-art models, skewing predictions towards the average of observed past values for the entity and providing worse-than-random predictive power in eccentric yet crucial cases; we name this phenomenon \textit{eccentricity bias}. We show that global error metrics like Root Mean Squared Error (RMSE) are insufficient to capture this bias, and we introduce Eccentricity-Area Under the Curve (EAUC) as a novel metric that can quantify it in all studied domains and models. We prove the intuitive interpretation of EAUC by experimenting with naive post-training bias corrections, and theorize other options to use EAUC to guide the construction of fair models. This work contributes a bias-aware evaluation of dyadic regression to prevent unfairness in critical real-world applications of such systems.
\end{abstract}


\section{Introduction}
\label{sec:introduction}

Dyadic data systems, which operate on data involving pairs of entities such as users and items, are ubiquitous in the modern world and significantly influence decisions made by individuals or which affect them. This work looks into three interrelated aspects: regression over dyadic data, the evaluation of such tasks, and the pervasive issue of unfairness biases in Artificial Intelligence (AI).

Dyadic data appear at the core of recommendation engines and countless other applications that involve understanding complex relationships between entities like users, products, or even drugs and job positions. Regression over dyadic data refers to the prediction of real values for a given pair of these entities, such as the expected trade volume between two companies. These predictions can influence anything, from purchasing decisions to employment opportunities, or medical treatment effectiveness. 

However, within these critical tasks, biases related to unfairness have profound implications, such as unequal access to opportunities or discriminatory decision-making processes \cite{caton2024fairness, tang2023and}. Regulations and guidelines are emerging globally to ensure fairness and ethics in AI systems to prevent these situations, such as the European Union's AI Act which regulates that AI systems must not be biased such they ``perpetuate historical patterns of discrimination" \cite{euaiact}, or the US AI Bill of Rights which will emphasize users ``should not face discrimination by algorithms and systems should be used and designed in an equitable way" \cite{hine2023blueprint}.

In the particular context of regression over dyadic data or \textit{dyadic regression}, unfairness biases have manifested, for example, as suboptimal recommendations that prioritize highly-rated items over more adequate ones for a user's needs, or amplify harmful messages in social media, favouring the creation of toxic echo chambers\cite{areeb2023filter, noordeh2020echo}.

One core challenge in evaluating dyadic regression tasks is the multi-dimensionality of such data. Typically, entities in the data are categorized into two sets and referred to as ``users'' and ``items''; we adopt this notation for the rest of this work. Because each observed value belongs to a particular pair of entities, each entity (user or item) can have a distinct local distribution of observed values. In this work, we prove such an additional layer of local value distributions in the data renders global error metrics like Root Mean Square Error (RMSE) or Mean Absolute Error (MAE) insufficient to assess the fairness aspect of dyadic regression models. We show that when these distributions are not uniform, RMSE-driven evaluation unfairly favours predictions biased towards the average past observed value of each user and item. This results in models that minimize RMSE but perform worse-than-random in non-regular cases that are often tied with fairness implications.

\begin{figure*}[h]
    \centering    \includegraphics[width=.8\textwidth]{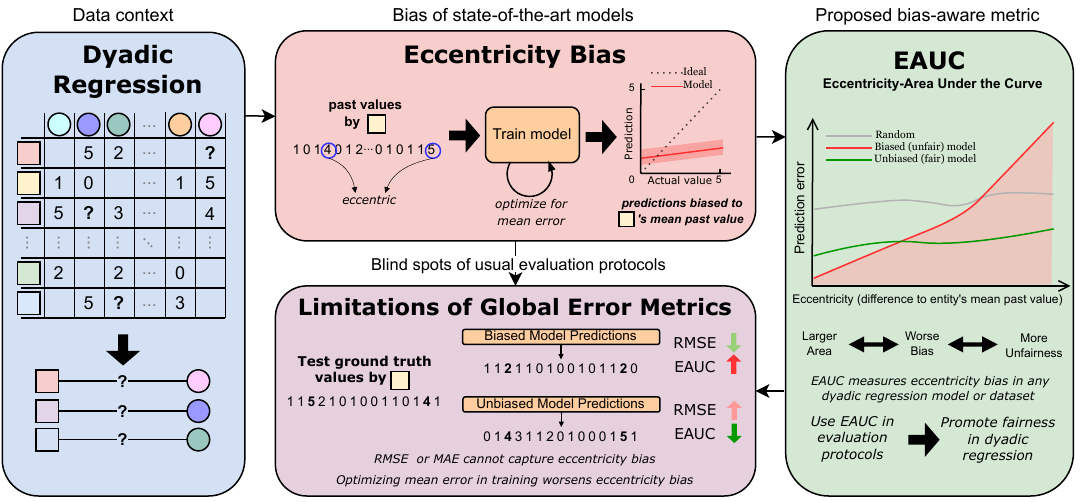}
    \caption{Main topics and contributions of this research work, including the definition of \textit{eccentricity bias} in dyadic regression, the limitations of global error metrics like RMSE or MAE for comprehensively evaluating said tasks, and the proposal of EAUC as a novel metric that can quantify the degree of eccentricity bias (and its subsequent unfairness) a dyadic regression model suffers.}
    \label{fig:summary}
\end{figure*}

In this research, we explore the limitations of basing the evaluation of dyadic regression tasks on global error metrics like RMSE, and the necessity of mitigating underlying unfairness in existing state-of-the-art models due to its negative discriminatory implications in critical fields. Our contributions data are summarized in Figure \ref{fig:summary} and include:

\begin{itemize}
    \item Unmasking that state-of-the-art models exhibit a strong predictive bias skewing predictions towards the average observed values of each user and item's past interactions (a phenomenon we name \textit{eccentricity bias}), leading to algorithmic unfairness and usage risks.
    \item Demonstrating how the classic global error metrics of the task, like RMSE and MAE, are insufficient to detect such a predictive bias in six real-world datasets from three fields with critical fairness implications (biomedicine, finances and product recommendation).
    \item Proposing a novel metric EAUC (Eccentricity-Area Under the Curve) which reliably provides a bias-oriented assessment of dyadic regression model performance in all of the aforementioned contexts.
    \item Demonstrating the intuitiveness of interpretation of EAUC scores, as well as proposing how they could be used to guide model construction, by experimenting with naive post-training bias corrections.
\end{itemize}
\section{Background}

This section formalizes dyadic data and dyadic regression tasks, overviews their classic evaluation methodology and its historical foundation, and demonstrates how we unmask the hidden \textit{eccentricity bias} in dyadic regression models that prove limitations in said evaluation protocols. 

\label{sec:background}
\subsection{Defining Dyadic Data and Dyadic Regression}
\label{sec:dyadicregression}

Dyadic data comprises interactions between pairs of entities, where each observation depicts a user interacting with an item. Typical tasks involve predicting a value given a pair of entities (e.g. predicting whether a job applicant and a vacant position are compatible) or predicting the optimal pair of entities given the task requirements (e.g. predicting the most suitable applicant for a job position).

Formally, let \(\mathcal{U}\) and \(\mathcal{I}\) be the set of users and items, respectively. A dyad is represented as \((u, i)\), where \(u \in \mathcal{U}\) and \(i \in \mathcal{I}\) are a user and an item, and the set of dyads constitute the dataset \(\mathcal{D} = \{(u, i) \mid u \in U, i \in I\}\).

This work focuses on dyadic regression tasks, which require predicting a real value for any dyad by determining the optimal regressor function $f: \mathcal{U} \times \mathcal{I} \rightarrow \mathbb{R}$ such that \(f\) predicts the optimal $\hat{r}_{ui} \in \mathbb{R}$ for dyad $(u, i)$. 

In the State of the Art, dyadic regression tasks appear often within Recommender Systems (RS), where dyadic regression can be used as a surrogate task to model Learning to Rank problems such as top-N recommendation, like ranking the best candidates for a job position, where fairness is a topic of paramount importance \cite{wang2023survey}. Regression on dyadic data also finds applications in various critical domains like finance, to estimate trade volume or pricing between stakeholders \cite{vidmer2015prediction} or biomedicine, to predict the adequate drug dosage required for a patient's treatment \cite{ota2022application, hu2018improvement}. As such, dyadic regression can directly affect people or their decisions, and predictions should be accurate, unbiased and fair due to their potential real-world impact.

\subsection{State-of-the-art evaluation: RMSE and Netflix Prize}
\label{sec:netflix}
Regarding evaluation, Root Mean Squared Error (RMSE) has long stood as the prevailing metric for dyadic regression tasks. Its adoption can be attributed to seminal events, notably the Netflix Prize \cite{bennett2007netflix}, a 2006 competition that challenged researchers to outperform the company's existing movie recommendation system Cinematc, attracting attention and innovation to the field \cite{bell2007bellkor, bell2007lessons, hallinan2016recommended}. RMSE was the official means of assessing algorithmic performance during the Netflix Prize competition, and is in essence a global error metric, formally defined as:

\begin{equation}
RMSE = \sqrt{\frac{1}{N}\sum_{k=1}^{N}(r_k - \hat{r_k})^2}
\end{equation}

\noindent where \(N\) is the total number of predictions, \(r_i\) is the observed value, and \(\hat{r_i}\) is the model's prediction. In the Netflix Prize, observed values were discrete ratings $r_{ui} \in {1, 2, 3, 4, 5}$, and models predicted real-valued ratings $\hat{r}_{ui} \in [1,5]$.

RMSE is simple to interpret (lower is better), facilitating its use. Other popular datasets like MovieLens \cite{harper2015movielens} further cemented RMSE and other global error metrics as the sole standard for evaluating dyadic regression tasks \cite{bobadilla2013recommender, chen2017performance}. However, they only offer an image of the overall error of the regressor,potentially neglecting biases in the predictions. 


\section{\hl{The need for EAUC: Limitations of RMSE in Dyadic regression tasks}}
\label{sec:rmselimitations}

\hl{In this section, we justify the proposal of our bias-aware evaluation metric EAUC by experimentally proving the limitations of global error metrics like RMSE in dyadic regression tasks. As traditionally used, these metrics evaluate over the dataset as a whole without distinguishing between different users, items, user-item pairs or their associated traits, and therefore we novelly postulate that they do not sufficiently capture the intricacies of dyadic data.

In the following, we reason and demonstrate the limitations of RMSE using the Netflix Prize data \mbox{\cite{bennett2007netflix}} and a Matrix Factorization (MF) model \mbox{\cite{rendle2020neural}}, a state-of-the-art dataset and model in the task, respectively.}

Table \ref{tab:rmse_per_rating} shows the Test RMSE for the examples of each unique rating (1 to 5) and the entire Test set. Although the system performs best for reviews with ratings 3 or 4 (the most common with $\sim28\%$ and $\sim33\%$ of ratings, respectively), predictions are always notably better than random, suggesting the absence of unfairness due to bias.

\begin{table}[ht]
\caption{RMSE of an MF model (overall and by rating)}
\centering
\resizebox{\columnwidth}{!}{
\begin{threeparttable}

\begin{tabular}{ccccccc}
\toprule
 & \textbf{Rating 1} & \textbf{Rating 2} & \textbf{Rating 3} & \textbf{Rating 4} & \textbf{Rating 5} & \textbf{All Ratings} \\
\midrule
\textbf{MF} & 1.681 & 1.082 & 0.636 & 0.594 & 0.914 & 0.822 \\
\textbf{Random} & 2.309 & 1.528 & 1.154 & 1.528 & 2.311 & 1.697 \\
\bottomrule 
\end{tabular}
\begin{tablenotes}
\item \textit{Values obtained by an MF model in the Netflix Prize dataset. The rating is the value $r_i$ to be predicted (here, a user's score to a given movie).}
\end{tablenotes}
\end{threeparttable}

}

\label{tab:rmse_per_rating}
\end{table}

However, dyadic data typically exhibit different rating distributions in different users or items. For instance, when rating movies, users may be less or more critical and movies may be less or more critically acclaimed. Consequently, this apparent lack of bias requires further examination. 

Let $\bar{r}_u$ or $\bar{r}_i$ be the average observed value for a user $u$ or an item $i$, such that:

\begin{equation}
\bar{r}_u = \frac{1}{|\mathcal{I}_u|}\sum_{i \in \mathcal{I}_u} r_{ui} \quad  
\bar{r}_i = \frac{1}{| \mathcal{U}_i|}\sum_{u \in \mathcal{U}_i} r_{ui} 
\end{equation}

\noindent where \(I_u\) and \(U_i\) are the set of items and users that \(u\) and \(i\) interacted with, respectively, and \(r_{ui}\) is an observed value.

Next, given a dyad $(u, i)$ pair, let the average of their $\bar{r}_u$ and $\bar{r}_i$ be their Dyadic Mean Value ($DMV_{ui}$). Formally:

\begin{equation}
DMV_{ui} = \frac{\bar{r_u} + \bar{r_i}}{2}
\end{equation}

Finally, let the eccentricity $Ecc_{ui}$ of an observed value $r_{ui}$ measure how much $r_{ui}$ deviates from the expected value for a dyad $(u, i)$ based on $u$'s and $i$'s past observations, as:

\begin{equation}
Ecc_{ui} = |r_{ui} - DMV_{ui}|
\end{equation}

\begin{figure}[h]
  \centering
  \includegraphics[width=.65\columnwidth]{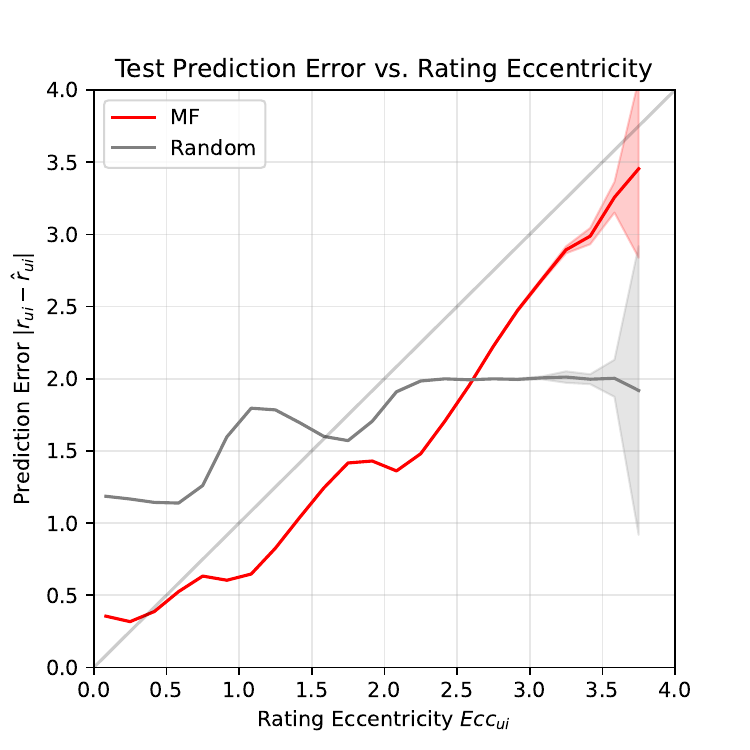}
  \caption{Eccentricity $Ecc_{ui}$ vs. prediction error $|\hat{r}_{ui} - r_{ui}|$ of an MF model and a uniform random predictor in Netflix Prize test examples (avg. and std. dev. of five runs). The $y=x$ line corresponds to predicting always the $DMV_{ui}$ of each example.}
  \label{fig:ecc_vs_mae_example}
\end{figure}

Using these concepts, Figure \ref{fig:ecc_vs_mae_example} displays the relationship between example eccentricity $Ecc_{ui}$ and prediction error $|\hat{r}_{ui} - r_{ui}|$ of the model and a uniform random predictor. We observe that, when the $r_{ui}$ to be predicted is close to the expected value $DMV_{ui}$ of the interaction (examples with low $Ecc_{ui}$), the model's predictions are very precise. However, the accuracy of predictions significantly deteriorates as $r_{ui}$ is further away from $DMV_{ui}$ (examples with high $Ecc_{ui}$). Worryingly, at high $Ecc_{ui}$, the MF model's predictions are considerably worse than predicting at random. 


\begin{figure}[!h]
  \centering
  \includegraphics[width=.8\columnwidth]
  {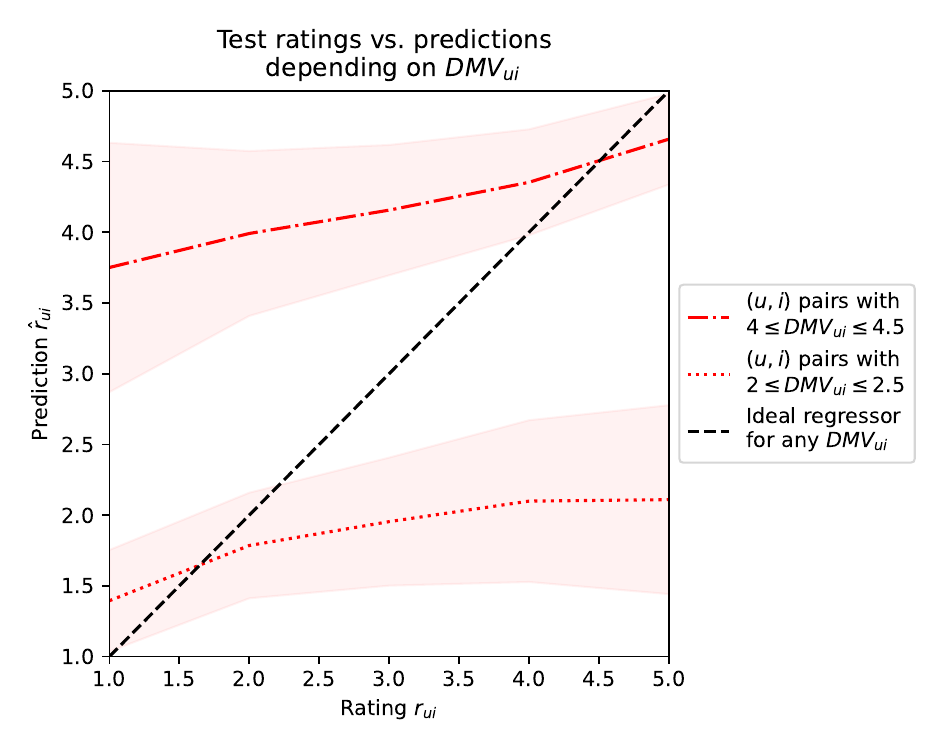}
  \caption{Observed vs. predicted values $r_{ui}$ and $\hat{r}_{ui}$ (avg. and std. dev.) of an MF model and an ideal dyadic regressor in Netflix Prize test examples, in two different scenarios of user-item pairs with low and high average ratings (low and high $DMV_{ui}$).}
  \label{fig:rating_vs_prediction_chart}
\end{figure}

To observe this condition from an alternative perspective, Figure \ref{fig:rating_vs_prediction_chart} shows the relation between ratings and predictions in two scenarios: test cases $(u,i)$ where $2.0 \leq DMV_{ui} \leq 2.5$ (user-item combinations with low average observed values), and test cases with $4.0 \leq DMV_{ui} \leq 4.5$ (user-item combinations with high average observed values). 

Although MF is mostly unbiased if viewing the dataset as a whole, in the low and high $DMV_{ui}$ scenarios we observe that the predictions are severely skewed towards the $DMV_{ui}$ of the dyad: the model assumes a critical user will not like a low-rated film even if they gave it a high rating, and vice-versa: we name this phenomenon as \textit{eccentricity bias}. As such, the knowledge generalization of the model is lower than an RMSE-guided evaluation would suggest: most of the global error minimization is achieved by blindly skewing predictions towards average observed values rather than by learning latent user and item features.

In dyadic regression applications with important fairness implications, eccentricity bias can cause unfairness and a potential risk of use. For instance, if predicting the adequate drug dosage of a specific pharmacological compound $i$ -which is typically given in large doses- for a given patient $u$ -who has a high resistance to treatments-, a prediction always close to $DMV_{ui}$ may result in severe overdosing of the patient and dangerous side effects; ultimately, patient safety would be compromised. 


\section{\hl{EAUC: Bias-aware Evaluation of Dyadic Regression}}
\label{sec:method}
\hl{This section formalizes EAUC (\textit{Eccentricity-Area Under The Curve}), our proposed novel alternative metric that can detect and quantify eccentricity bias and its subsequent unfairness in dyadic regression models' prediction. In the following, we discuss both its theoretical foundation and its practical usage in evaluation policies.}

\subsection{\hl{Theoretical derivation of EAUC}}

\hl{The underlying theory of EAUC is to, by accounting for the eccentricity of examples, ensure that the measured performance of a model can reflect its ability to handle cases that deviate significantly from the average past values $DMV_{ui}$ of the entities involved; these cases often represent critical edge cases where fairness implications are highest, but they are neglected by global error metrics like RMSE.

As codified in Algorithm \mbox{\ref{alg:eauc}}, EAUC is computed as follows:}

\begin{enumerate}
    \item Calculate in the Train set the mean observed values $\bar{r}_u$ and $\bar{r}_i$ for each user $u$ or item $i$.
    \item For each test example $j : (u,i,r_{ui},\hat{r}_{ui})$, calculate:
    \begin{itemize}
        \item $Ecc_j$, the eccentricity of the observed value $r_{ui}$.
        \item $\epsilon_j$, the absolute prediction error $|\hat{r}_{ui} - r_{ui}|$.
    \end{itemize}
    \item Sort Test examples into an ordered sequence of pairs $(Ecc_j, \epsilon_j)$ with increasing $Ecc_j$ values.
    \item Integrate to determine the area $A$ under the curve defined by eccentricities $Ecc$ and absolute errors $\epsilon$.
    \item Normalize the area $A$ within the range [0, 1] by dividing by the area of a rectangle with sides $Ecc_{max}$ and $\epsilon_{max}$ (both can be computed as $max(r_{ui}) - min(r_{ui})$).
\end{enumerate}

\begin{algorithm}
\caption{ \begin{footnotesize}
EAUC (Eccentricity-Area Under the Curve)
\end{footnotesize}}
\begin{algorithmic}[5]
\State \textbf{Inputs:} Train and test examples $D_{\text{train}} = \{(u, i, r_{ui})\}$, $D_{\text{test}} =\{(u, i, r_{ui})\}$

\State \textbf{Outputs:} Calculated score EAUC
\State Initialize $\bar{r}_{users} \leftarrow \mathbb{R}^{|\mathcal{U}|}$, $\bar{r}_{items} \leftarrow \mathbb{R}^{|\mathcal{I}|}$ 
\For{$j = 1$ to $|\mathcal{U}|$}
\State $\bar{r}_{users_j} \leftarrow \frac{1}{|D_{\text{train}_j}|} \sum_{(u,i,r_{ui}) \in D_{\text{train}_j}} r_{ui}$
\EndFor
\For{$j = 1$ to $|\mathcal{I}|$}
\State $\bar{r}_{items_j} \leftarrow \frac{1}{|D_{\text{train}_j}|} \sum_{(u,i,r_{ui}) \in D_{\text{train}_j}} r_{ui}$
\EndFor
\State Initialize $Ecc \leftarrow \mathbb{R}^{|\mathcal{D}_{\text{test}}|}$  and $\epsilon \leftarrow \mathbb{R}^{|\mathcal{D}_{\text{test}}|}$
\For{$j = 1$ to $|D_{\text{test}}|$}
\State $(u,i,r_{ui},\hat{r}_{ui}) \leftarrow D_{\text{test}_j}$ 
\State $Ecc_j \leftarrow |r_{ui} - \frac{1}{2} \cdot (\bar{r}_{users_u} + \bar{r}_{items_i})|$
\State $\epsilon_j \leftarrow |\hat{r}_{ui} -  r_{ui}|$
\EndFor
\State Sort $(Ecc_j, \epsilon_j)$  pairs by ascending $Ecc_j$
\State Initialize $A \leftarrow 0$
\For{$j = 2$ to $|D_{\text{test}}|$}
\State $\Delta A_j \leftarrow \frac{1}{2} \cdot (Ecc_j - Ecc_{j-1}) \cdot (\epsilon_j + \epsilon_{j-1})$
\State $A \leftarrow A + \Delta A_j$
\EndFor
\State EAUC $\leftarrow \frac{A}{(max(r_{ui}) - min(r_{ui}))^2} ; (u, i, r_{ui}, \hat{r}_{ui}) \in D_{\text{test}}$
\end{algorithmic}
\label{alg:eauc}
\end{algorithm}

The EAUC of a given model $f : \mathcal{U} \times \mathcal{I} \rightarrow \mathbb{R}$ in a given dataset $\mathcal{D}$ with training and test subsets $\mathcal{D}_{train}$ and $\mathcal{D}_{test}$ may also be articulated as a differentiable function $EAUC(f, \mathcal{D})$ in formulaic fashion, such that: 
\begin{flalign}
EAUC(f, \mathcal{D}) = \frac{\sum\limits_{x_i\in\mathcal{D}_{test}}\frac{(Ecc_{x_i} -Ecc_{x_{i-1}})*(\epsilon_{x_i} + \epsilon_{x_{i-1}})}{2}}{(max(r_{ui})-min(r_{ui}))^2}\quad
\end{flalign}
\noindent
\begin{align*}
\text{Where}\quad\quad\quad Ecc_{\{u, i, r_{ui}\}} =& |r_{ui}- \frac{1}{2}(\bar{r}_u+\bar{r}_i)|, \nonumber&&\\
\epsilon_{\{u, i, r_{ui}\}} =& (r_{ui}-f(u,i))^2, \nonumber&&\\
\bar{r}_u = &\frac{1}{|\mathcal{I}_u|}\sum_{i \in \mathcal{I}_u \subseteq \mathcal{I}_{\mathcal{D}_{train}}} r_{ui},\nonumber  &&\\
\bar{r}_i= &\frac{1}{| \mathcal{U}_i|}\sum_{u \in \mathcal{U}_i \subseteq \mathcal{U}_{\mathcal{D}_{train}}} r_{ui},\nonumber  &&\\
x_i \in &\{u, i, r_{ui}\};\nonumber &&\\
\nonumber\\
\text{With}\quad(Ecc_{x_i} - Ecc_{x_{i-1}})\geq&0 &&
\label{eqn:eauc}
\end{align*}

\subsection{\hl{Usage of EAUC for model evaluation}}
\label{EAUC:discussion}
EAUC produces a value in the range [0, 1], where lower values indicate lower eccentricity bias and better predictive power. An ideal and unbiased model with perfect predictions achieves $EAUC=0$, while a naive predictor that always outputs $DMV_{ui}$ achieves $EAUC=0.5$ (if measured on examples with eccentricities spanning $[0, max(r_{ui})]$).

\begin{figure}[htbp]
    \centering
    \includegraphics[width=\columnwidth]{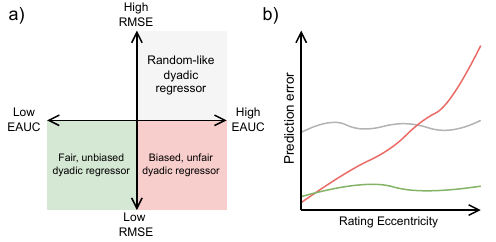}
    \caption{Usage of EAUC for evaluation of dyadic regression tasks, by a) characterizing the EAUC vs. RMSE trade-off to incorporate the fairness component into model evaluation, and b) analyzing the computed Eccentricity vs. Prediction Error curve  (color codes in Subfigure b) represent the corresponding model characteristics in Subfigure a)). While a dyadic regressor with high RMSE and low EAUC is plausible in theory, it is unlikely in practice as models minimize global errors during training.}
    \label{fig:EAUCinterpretation}
\end{figure}

During the evaluation of dyadic regression tasks, there exist two intuitive approaches to using EAUC as an informative analytic descriptor of bias-aware performance and model fairness, as seen in Figure \mbox{\ref{fig:EAUCinterpretation}}: 

\begin{enumerate}[label=(\alph*)]
    \item Characterizing real model performance by incorporating EAUC as a bias-aware evaluation dimension to classic metrics like RMSE. For instance, a model is comparatively better than another if it achieves not only lower RMSE but also lower EAUC. Similarly, a model with low RMSE may be inadequate for tasks with strict fairness requirements if it has high EAUC.
    \item Inspecting the Eccentricity vs. Error curve which allows, for instance, to observe whether a trained dyadic regressor has worse-than-random performance on critical eccentric cases, causing unfair predictions.

\end{enumerate}

Consequently, EAUC offers a valuable complement to traditional metrics like RMSE and MAE, which quantify average errors but neglect predictive bias learnt by models to optimise their predictions in dyadic settings. 

Moreover, since eccentricity bias leads to discriminatory outcomes when fairness is involved, EAUC also behaves as measure of \textit{fairness} in dyadic regression tasks, being able to complement and synergize with classic fairness metrics. In this aspect, it is plausible that EAUC will become a reliable detector of an -until now- unknown source of unfairness in this type of task, i.e. eccentricity bias.

As an indicator of unfairness in dyadic regression, EAUC may be used to directly detect the presence of eccentricity-based individual or group unfairness \mbox{\cite{pessach2022review}}, which can later be characterized using utility or parity-based fairness metrics \mbox{\cite{yao2017beyond}}. Conversely, the inverse is also possible: while most group or individual unfairness are often attributed to representational or class-stereotypical biases in the data \mbox{\cite{dominguez2024less}}, they often cannot fully explain the source of fairness\mbox{\cite{9808125}}; EAUC and its analysis may be used to better characterize the source of otherwise opaque unfairness phenomenons. One final property of EUAC is its ability to, unlike classic individual and group-based metrics modelling parity and utility \mbox{\cite{castelnovo2021zoo}}, detect eccentricity-based unfairness that affects all entities (e.g. users) of the population as a whole, rather than only specific subgroups or individuals. As a result, EAUC can improve our ability to both detect and better characterize new and known unfairness in dyadic regression.

\subsection{Measuring Dataset Difficulty for Eccentricity Bias}

Using the same dyadic regression dataset, EAUC can be used to measure and compare the eccentricity bias of different models. Between datasets, however, a difference in EAUC could be caused not only by model adequateness but also due to the characteristics or difficulty of the dataset. Dataset difficulty metrics are used in Machine Learning to assess their upfront difficulty and contextualize model performance \cite{li2014assessing, collins2018evolutionary, ho2002complexity}. In dyadic data, dataset sparsity and size are used to measure local and global quantities of information, but do not characterize the local distributions of observed values of individual entities, which we proved influence the learning of models. 

We could argue that, as state-of-the-art models learn by minimizing a global error metric like MSE, their eccentricity bias can be more pronounced if the observed values for each entity are concentrated around its average rating. Imagine a user $u$'s observed values follow a normal distribution $X \sim \mathcal{N}(\bar{r}_{u},\,\sigma^{2})$, with mean $\bar{r}_{u}$ and standard deviation $\sigma$. The lower $\sigma$ is, the larger the reward for biasing predictions towards $\bar{r}_{u}$ instead of learning latent general knowledge. Conversely, if $u$'s observed values are uniformly distributed, the model must attain this real latent knowledge about users and items to minimize prediction errors.

Based on this reasoning, we propose adapting the Kolmogorov-Smirnov uniformity statistic $D_{KS}$ to approximate the severity of the eccentricity biases induced by a given dataset in models learnt with global error optimization. For a given dataset $\mathcal{D}$, we compute this entity-wise uniformity statistic, which we term $D_{KS_{\mathcal{D}}}$, as:

\begin{equation}
D_{KS_{\mathcal{D}}} = \frac{ \sum_{u \in |\mathcal{U}|} D_{KS_{u}}+\sum_{i \in |\mathcal{I}|} D_{KS_{i}}}{|\mathcal{U}| + |\mathcal{I}|}
\end{equation}

\noindent where $D_{KS_{u}}$ or $D_{KS_{i}}$ is the Kolmogorov-Smirnov statistic of a given user's or item's distribution of observed values, measuring its maximum deviation to a uniform distribution $X\sim\mathcal{U}(min(r_{ui}),max(r_{ui}))$. Lower $D_{KS_{\mathcal{D}}}$ correspond to more uniform observed value distributions across the dataset's users and items, where models are less tempted to naively predict the $DMV_{ui}$ for any $(u, i)$ dyad. Conversely, higher $D_{KS_{\mathcal{D}}}$ indicate non-uniform distributions, luring models into biasing predictions towards the respective $DMV_{ui}$, causing eccentricity bias and unfairness.

\section{Experimental Setup}
\label{sec:setup}

In this section, we cover different aspects of our experimental design, including datasets, models and other implementation details, with the objective of analyzing both the extent of eccentricity bias in real-world dyadic regression contexts and the models, and the efficacy of EAUC and inefficacy of RMSE to detect such a bias. 

\subsection{Datasets}
\label{sec:datasets}

We selected for our experiments dyadic regression datasets in three contexts with significant fairness implications: biomedicine and pharmacology, product recommendation, and finances and lending. 

Table \mbox{\ref{tab:datasets} }shows basic statistics of the used datasets: Netflix Prize \mbox{\cite{bennett2007netflix}} and Movielens 1M \mbox{\cite{harper2015movielens}} are product recommendation datasets, where the objective is to predict the rating a given user will give to a movie after viewing it. GDSC1 \mbox{\cite{yang2012genomics}} and CTRPv2 \mbox{\cite{basu2013interactive}} contain biomedical pharmacology data, where the goal is to predict the effectiveness of a given drug in fighting against specific cancer cell lines. IMF DOTS 2023 \mbox{\cite{imfdots}} and \textit{Kiva} \mbox{\cite{kivaml17}} contain financial interaction data; the former seeks to predict the potential Value of Trade between two economic entities (in this case, UN countries or regions), and the later consists in predicting the normalized lent amount by a lender to a specific cluster of microloans.

\begin{table}[h!]

\centering

\caption{Basic statistics of the used datasets}
\label{tab:datasets}
\renewcommand{\arraystretch}{1.5}
\resizebox{\linewidth}{!}%
{\begin{threeparttable}
\begin{tabular}{llrrrr}
\hline
\textbf{Domain} &
  \textbf{Dataset} &
  \multicolumn{1}{l}{\textbf{Examples}} &
  \multicolumn{1}{l}{\textbf{Users}} &
  \multicolumn{1}{l}{\textbf{Items}} &
  \multicolumn{1}{l}{\textbf{Density}} \\ \hline
\multirow{2}{*}{\begin{tabular}[c]{@{}l@{}}Product\\ Recommendation\end{tabular}}        & Movielens 1M    & 1,000,209   & 6,040   & 3,706  & 4.5\%  \\
                                                                                      & Netflix Prize   & 100,480,507 & 480,189 & 17,770 & 1.2\%  \\ \cline{1-1}
\multirow{2}{*}{\begin{tabular}[c]{@{}l@{}}Biomedical \&\\ Pharmacology\end{tabular}} & GDSC1           & 99,948      & 969     & 402   & 25.6\% \\
                                                                                      & CTRPv2          & 108,224     & 829     & 544   & 24.0\% \\ \cline{1-1}
\multirow{2}{*}{\begin{tabular}[c]{@{}l@{}}Financial \&\\ Lending\end{tabular}}       & IMF DOTS 2023   & 16,776      & 233     & 233   & 32.0\% \\
                                                                                      & Kiva Microloans & 110,371     & 4,005   & 2,673  & 1.0\%  \\ \hline
\end{tabular}
\begin{tablenotes}
\item \textit{Density indicates quantity of implicit feedback information available per entity (user or item) as $|\mathcal{D}| / (|\mathcal{U}| \times |\mathcal{I}|)$}
\end{tablenotes}
\end{threeparttable}
}
\end{table}

We applied 90/10 random train/test data splitting to all datasets, and used only implicit feedback information (no extra features). For GDSC1, CTRPv2 and IMF DOTS 2023, we sampled  $1/3$ of the data before performing the train-test split, to reproduce a higher sparsity more akin to real-world use-case scenarios of dyadic regression.
    
\subsection{Models}
\label{sec:models}

We consider a diverse array of state-of-the-art dyadic regression models conformed by Bayesian SVD++ \cite{rendle2019difficulty} (matrix decomposition), Glocal-K \cite{han2021glocal} (auto-encoders and convolution kernels), GC-MC \cite{berg2017graph} (graph auto-encoders), and MF \cite{rendle2020neural} (simple matrix factorization). We also use two naive predictors as baselines: a random uniform predictor (Random) and a predictor that always outputs the $DMV_{ui}$ of the example (Dyad Average). This variety of algorithmic approaches to allows for a comprehensive evaluation of the prevalence of eccentricity bias, the limitations of global error metrics like RMSE and MAE, and the ability of EAUC to quantify its presence, absence or severity.

\subsection{Experimental Implementation Details}

For each model and dataset, we use the hyperparameters for each dataset described in the model's original publication or code repository, to respect reproducibility; if such configuration is not present, we fall back on those for the dataset with most similar size. Appendix \ref{apx:hyperparameters} lists the full relation of used hyperparameters. 

Model training and evaluation is done on a dedicated machine with 16GB RAM, an NVIDIA GeForce RTX 2060 Super GPU, and an Intel Core i7-10700K CPU @ 3.80GHz.

Our evaluation framework is published in a public code repository\footnote{https://github.com/Kominaru/SUB-COFI} for reproducibility, where we also make publicly available an efficient implementation of EAUC for use in future research in fairness-aware dyadic regression.

\section{Results and Discussion}
\label{sec:results}

The results shown in Table \ref{tab:resultsmetrics} include the global error performance (RMSE, MAE) and bias-oriented performance (our novel metric EAUC) obtained for each dataset and state-of-the-art model or baseline. 

\begin{table*}[h]

\caption{Global error RMSE and MAE and eccentricity bias EAUC per model and dataset, and ``fairness difficulty'' $D_{KS_{\mathcal{D}}}$ per dataset.}
\label{tab:resultsmetrics}
\centerline{\textbf{Product Recommendation}}
\vspace{0.1cm}
\resizebox{\textwidth}{!}{%
\begin{tabular}{lrrrlrrr}
\toprule
 &
  \multicolumn{2}{l}{\textbf{Movielens 1M}} &
  \multicolumn{1}{l}{$D_{KS_{\mathcal{D}}}$: 0.412} &
   &
  \multicolumn{2}{l}{\textbf{Netflix Prize}} &
  \multicolumn{1}{l}{$D_{KS_{\mathcal{D}}}$: 0.429} \\ \cmidrule(lr){2-4} \cmidrule(l){6-8} 
 &
  \multicolumn{1}{l}{\textbf{RMSE}} &
  \multicolumn{1}{l}{\textbf{MAE}} &
  \multicolumn{1}{l}{\textbf{EAUC}} &
   &
  \multicolumn{1}{l}{\textbf{RMSE}} &
  \multicolumn{1}{l}{\textbf{MAE}} &
  \multicolumn{1}{l}{\textbf{EAUC}} \\ \midrule
Random &
  1.710 ± 0.004 &
  1.397 ± 0.004 &
  0.361 ± 0.016 &
   &
  \multicolumn{1}{r}{1.696 ± 0.000} &
  \multicolumn{1}{r}{1.386 ± 0.001} &
  \multicolumn{1}{r}{{\ul 0.395 ± 0.004}} \\
Dyad Average &
  0.956 ± 0.002 &
  0.770 ± 0.002 &
  0.370 ± 0.013 &
   &
  \multicolumn{1}{r}{{\ul 0.964 ± 0.000}} &
  \multicolumn{1}{r}{{\ul 0.781 ± 0.000}} &
  \multicolumn{1}{r}{0.435 ± 0.008} \\ \cmidrule(r){1-4} \cmidrule(l){6-8} 
Bayesian SVD++ &
  {\ul 0.830 ± 0.003} &
  {\ul 0.650 ± 0.002} &
  \textbf{0.297 ± 0.013} &
   &
  - &
  - &
  - \\
Glocal-K &
  \textbf{0.826 ± 0.003} &
  \textbf{0.643 ± 0.002} &
  {\ul 0.297 ± 0.029} &
   &
  - &
  - &
  - \\
GC-MC &
  0.846 ± 0.005 &
  0.665 ± 0.004 &
  0.323 ± 0.028 &
   &
  - &
  - &
  - \\
MF &
  0.840 ± 0.003 &
  0.657 ± 0.003 &
  0.315 ± 0.013 &
   &
  \multicolumn{1}{r}{\textbf{0.826 ± 0.001}} &
  \multicolumn{1}{r}{\textbf{0.635 ± 0.001}} &
  \multicolumn{1}{r}{\textbf{0.354 ± 0.006}} \\ \bottomrule
\end{tabular}
}
\vspace{0.2cm} \\
\centerline{\textbf{Biomedicine \& Pharmacology}}
\vspace{0.1cm}
\resizebox{\textwidth}{!}{%
\begin{tabular}{lrrrlrrr}
\toprule
 &
  \multicolumn{2}{l}{\textbf{GDSC1}} &
  \multicolumn{1}{l}{$D_{KS_{\mathcal{D}}}$: 0.384} &
   &
  \multicolumn{2}{l}{\textbf{CTRPv2}} &
  \multicolumn{1}{l}{$D_{KS_{\mathcal{D}}}$: 0.390} \\ \cmidrule(lr){2-4} \cmidrule(l){6-8} 
 &
  \multicolumn{1}{l}{\textbf{RMSE}} &
  \multicolumn{1}{l}{\textbf{MAE}} &
  \multicolumn{1}{l}{\textbf{EAUC}} &
   &
  \multicolumn{1}{l}{\textbf{RMSE}} &
  \multicolumn{1}{l}{\textbf{MAE}} &
  \multicolumn{1}{l}{\textbf{EAUC}} \\ \midrule
Random         & 6.964 ± 0.069          & 5.864 ± 0.066          & 0.134 ± 0.015          &  & 5.804 ± 2.297          & 5.689 ± 0.072          & 0.207 ± 0.036          \\
Dyad Average   & 1.672 ± 0.016          & 1.301 ± 0.010          & 0.081 ± 0.008          &  & 1.841 ± 0.011          & 1.332 ± 0.007          & 0.179 ± 0.016          \\ \cmidrule(r){1-4} \cmidrule(l){6-8} 
Bayesian SVD++ & \textbf{1.029 ± 0.035} & \textbf{0.747 ± 0.007} & \textbf{0.039 ± 0.006} &  & 1.244 ± 0.010          & {\ul 0.865 ± 0.006}    & \textbf{0.074 ± 0.011} \\
Glocal-K       & 1.063 ± 0.020          & 0.780 ± 0.015          & 0.048 ± 0.007          &  & {\ul 1.241 ± 0.014}    & 0.866 ± 0.007          & 0.086 ± 0.014          \\
GC-MC          & {\ul 1.041 ± 0.008}    & {\ul 0.754 ± 0.007}    & 0.047 ± 0.005          &  & \textbf{1.220 ± 0.019} & \textbf{0.841 ± 0.011} & {\ul 0.078 ± 0.020}    \\
MF             & 1.074 ± 0.015          & 0.795 ± 0.011          & {\ul 0.042 ± 0.005}    &  & 1.279 ± 0.020          & 0.902 ± 0.006          & 0.105 ± 0.012          \\ \bottomrule
\end{tabular}
}
\vspace{0.2cm} \\
\centerline{\textbf{Financial \& Lending}}
\vspace{0.1cm}
\begin{threeparttable}
\resizebox{\textwidth}{!}{%
\begin{tabular}{lrrrlrrr}
\toprule
 &
  \multicolumn{2}{l}{\textbf{IMF DOTS 2023}} &
  \multicolumn{1}{l}{$D_{KS_{\mathcal{D}}}$: 0.3717} &
   &
  \multicolumn{2}{l}{\textbf{Kiva Microloans}} &
  \multicolumn{1}{l}{$D_{KS_{\mathcal{D}}}$: 0.873} \\ \cmidrule(lr){2-4} \cmidrule(l){6-8} 
 &
  \multicolumn{1}{l}{\textbf{RMSE}} &
  \multicolumn{1}{l}{\textbf{MAE}} &
  \multicolumn{1}{l}{\textbf{EAUC}} &
   &
  \multicolumn{1}{l}{\textbf{RMSE}} &
  \multicolumn{1}{l}{\textbf{MAE}} &
  \multicolumn{1}{l}{\textbf{EAUC}} \\ \midrule
Random         & 4.044 ± 0.102          & 3.350 ± 0.104          & 0.189 ± 0.018          &  & 1.111 ± 0.003          & 0.952 ± 0.004          & \textbf{0.408 ± 0.006} \\
Dyad Average   & 1.355 ± 0.033          & 1.076 ± 0.034          & 0.136 ± 0.019          &  & {\ul 0.413 ± 0.009}    & 0.227 ± 0.004          & 0.498 ± 0.003          \\ \cmidrule(r){1-4} \cmidrule(l){6-8} 
Bayesian SVD++ & {\ul 0.937 ± 0.018}    & 0.664 ± 0.013          & 0.080 ± 0.024          &  & 0.472 ± 0.007          & 0.258 ± 0.003          & 0.461 ± 0.011          \\
Glocal-K       & \textbf{0.889 ± 0.013} & \textbf{0.640 ± 0.014} & \textbf{0.075 ± 0.015} &  & 0.415 ± 0.012          & {\ul 0.213 ± 0.003}    & 0.461 ± 0.020          \\
GC-MC          & 0.946 ± 0.049          & {\ul 0.658 ± 0.029}    & {\ul 0.080 ± 0.021}    &  & 0.414 ± 0.009          & 0.226 ± 0.009          & 0.469 ± 0.015          \\
MF             & 1.005 ± 0.012          & 2.111 ± 0.046          & 0.100 ± 0.017          &  & \textbf{0.405 ± 0.009} & \textbf{0.205 ± 0.005} & {\ul 0.440 ± 0.007}    \\ \bottomrule
\end{tabular}
}
\begin{tablenotes}
\item \textit{In each dataset, the first group of results  are naive baselines, and the second group are state-of-the-art models (avg. and std. dev. of 5 runs). Bold and underlined values represent the best and second best-performing model, respectively, of each dataset and metric. On Netflix Prize, ML models other than MF could not be executed due to dataset size and model computational complexity.}
\end{tablenotes}
\end{threeparttable}
\end{table*}

It can be observed that that all studied models suffer from eccentricity bias in varying degrees, yet a lower RMSE does not correlate with a lower bias (and therefore, a lower EAUC); for instance, on CTRPv2 (drug efficacy prediction), Bayesian SVD++ holds a much lower bias (EAUC) than other ML models despite having a significantly higher RMSE. Similarly, the non-uniformity of entities' value distributions on Kiva (microloan lending prediction) leads to extreme eccentricity biases across all models, with EAUC scores surpassing that of random baselines, despite showing apparently lower global error through RMSE and MAE; in a context with such strict societal impact, this inevitably leads to unfair and discriminatory outcomes that otherwise can not be measured if using only RMSE and/or MAE.

\begin{figure*}[!h]
\centering    
\begin{minipage}{0.32\textwidth}
    \centering
    \textbf{Product Recommendation}
    \includegraphics[width=.75\linewidth]
    {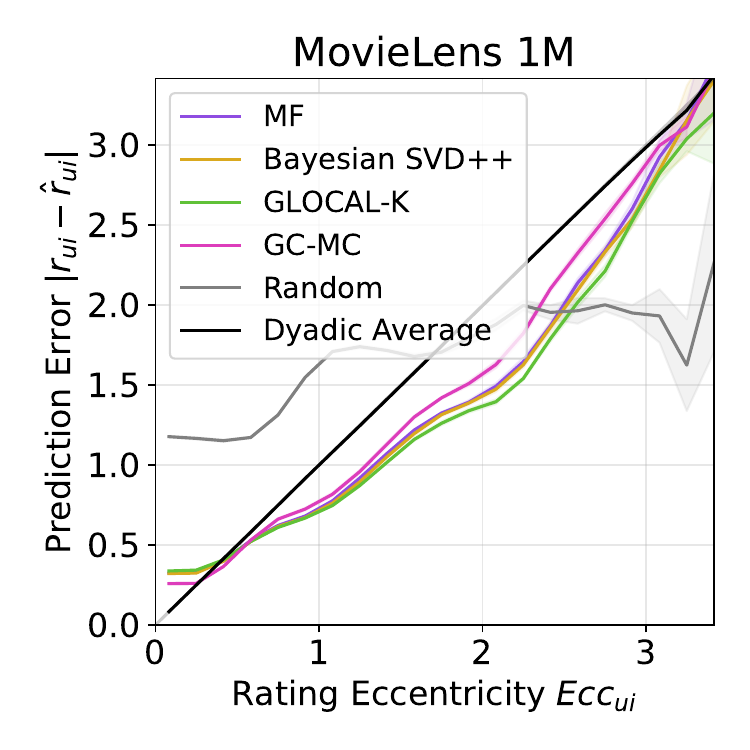}
    \includegraphics[width=.75\linewidth]{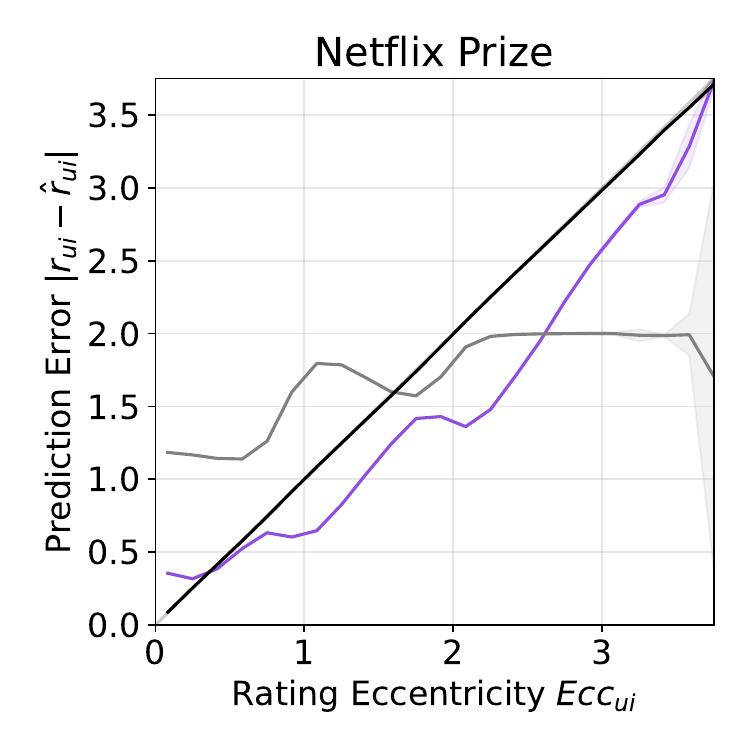}
\end{minipage}
\begin{minipage}{0.32\textwidth}
    \centering
    \textbf{Biomedicine \& Pharmacology }
    \includegraphics[width=.75\linewidth]{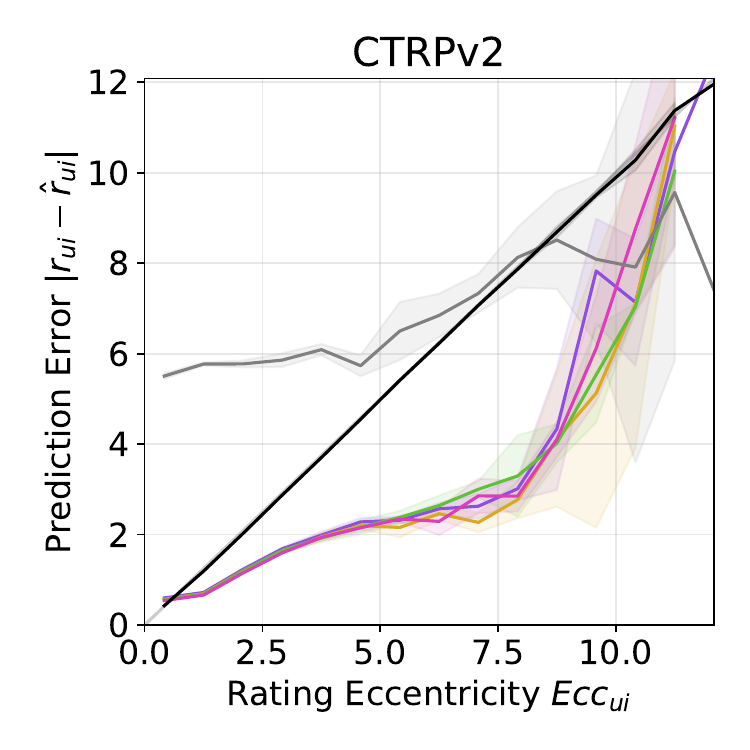}
    \includegraphics[width=.75\linewidth]{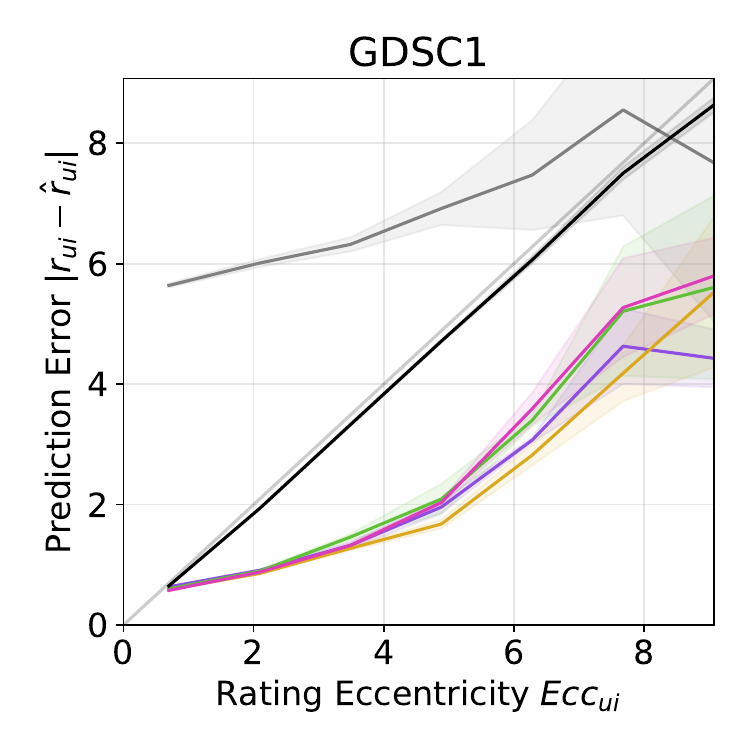}
    
\end{minipage}
\begin{minipage}{0.32\textwidth}
    \centering
    \textbf{Financial \& Lending}
    \includegraphics[width=.75\linewidth]{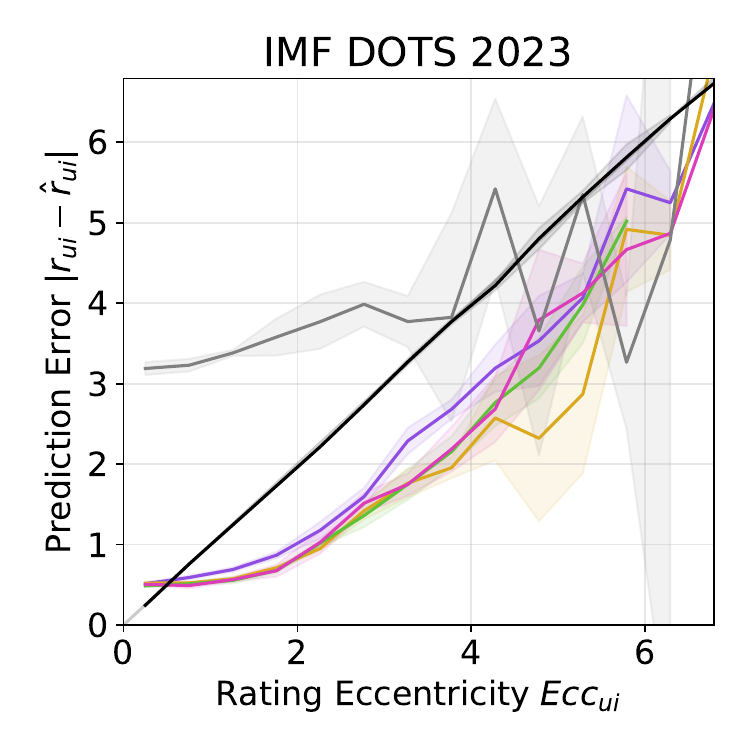}
    \includegraphics[width=.75\linewidth]{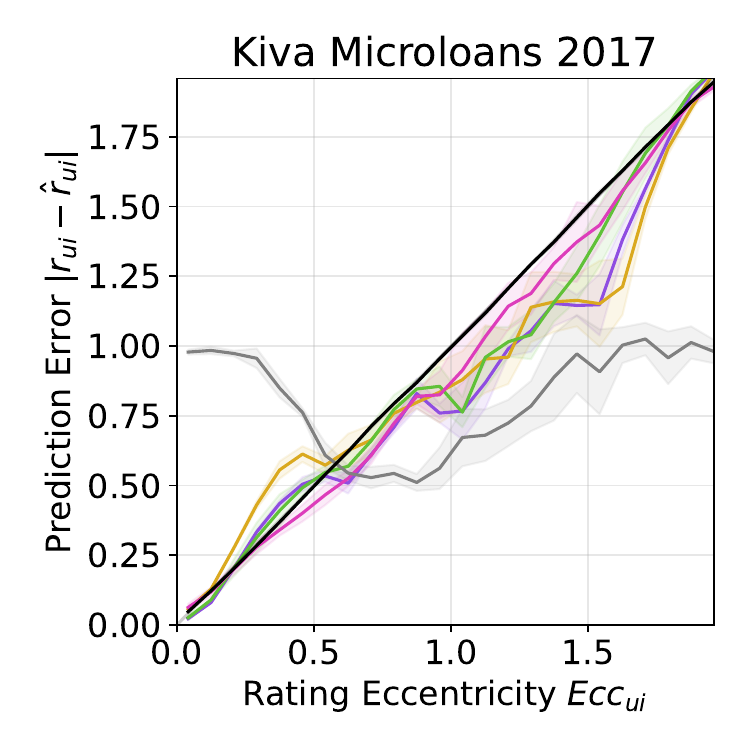}
\end{minipage}

\caption{Relationship between eccentricity of observed values and prediction error (lower is better) for each model and dataset (avg. and std. dev. of 5 runs). On Netflix Prize, ML models other than MF could not be executed due to dataset size and model computational complexity.}
\label{fig:ecc_vs_error_graphs}
\end{figure*}

Figure \mbox{\ref{fig:ecc_vs_error_graphs}} shows for each dataset the relation between the eccentricity of observed values and the prediction errors. Model error is near-zero when the eccentricity is very low but increases almost linearly with eccentricity. This is particularly severe in datasets with very non-uniform observed value distributions, such as Netflix and Kiva, whereglobal optimizers will be rewarded for always predicting values near $DMV_{ui}$. Indeed, it can also be observed in Table \mbox{\ref{tab:resultsmetrics}} that the $D_{KS_{\mathcal{D}}}$ for a dataset correlates with the measured EAUC,  indicating that less uniform local distributions of observed values of users and items lead to learnt models with more severe eccentricity biases.

\begin{figure*}[!h] 
    \centering
    \includegraphics[width=\textwidth]{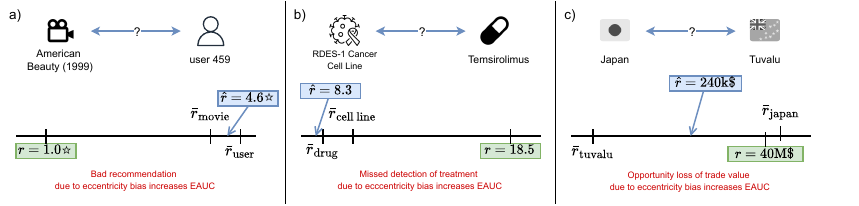}
    \caption{Examples of EAUC detecting eccentricity bias and unfairness of an MF model in three contexts with strict fairness requirements: product recommendation (a) on MovieLens, prediction of drug effectiveness on cancer variants (b) on CTRPv2, and prediction of international value of trade (c) on IMF DOTS. In all cases, the model's prediction $\hat{r}$ is biased towards the average of the mean observed values $\bar{r}$ of the two entities, causing a higher error compared to the actual label $r$ on an eccentric example.}
    \label{fig:eccexamples}
\end{figure*}

Figure \mbox{\ref{fig:eccexamples}} observably showcases the ability of EAUC to detect individual unfair outcomes caused by the eccentricity bias of dyadic regression models in three contexts with strict fairness requirements: product recommendation (Subfigure a), drug response prediction (Subfigure b) and trade revenue prediction (Subfigure c). When eccentric cases hold significant importance, such as in the aforementioned contexts, avoiding large errors in eccentric cases is more critical for system usability and fairness than seeking near-zero errors in common, predictable scenarios.

Summarizing, we have proven the evaluation of dyadic regression should not be limited to global error metrics like RMSE; it should also consider the evaluation of entity-wise biases, which we achieve with our novel metric EAUC.

\subsection{Usage of EAUC to mitigate eccentricity bias}
\label{sec:corrs}

\begin{figure}[htbp]
    \centering
\includegraphics[width=.9\columnwidth]{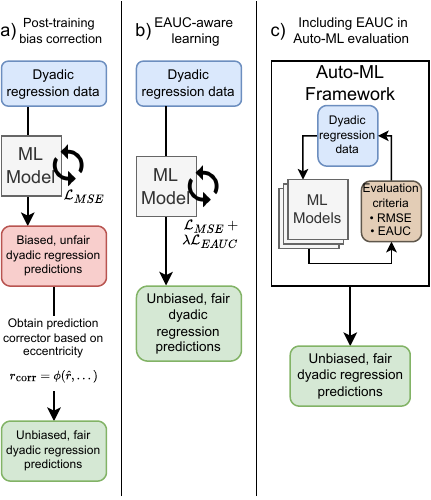}
    \caption{Potential usages of EAUC and eccentricity bias to guide fair model construction through a) post-training bias correction, b) EAUC-aware learning, or c) integration of EAUC into Auto-ML pipelines.}
    \label{fig:eauc_usage}
\end{figure}

Until now, we proved EAUC can be used as a reliable metric to detect and quantify eccentricity bias and contribute to a fairness-aware evaluation of dyadic regression; we theorize that EAUC can also guide the obtaining of fair dyadic regression models which mitigate or avoid eccentricity bias. While the development of such models is out of the scope of this article, Figure \mbox{\ref{fig:eauc_usage}} suggests some potential avenues for fairness-oriented training and selection:

\begin{enumerate}[label=(\alph*)]
    \item One naive option is to derive eccentricity-based correction functions to de-bias the unfair predictions of biased trained models. This is, for a biased dyadic regression model $f$ we could define a correcting $\phi$ that remaps the outputs of $f$ such that:
    \begin{align}
        \hat{r_{ui}}_{corr} = \phi(f(u,i), \dots)\nonumber\\ 
        EAUC(\phi \circ f, \mathcal{D}) < EAUC(f,  \mathcal{D})
    \end{align}
    Post-training bias-correction methods have the benefit of taking advantage of the already trained model, avoiding the computational complexity of training a fair model from scratch (as the correction function $\phi$ will be typically simpler). In fairness contexts, a similar approach has been previously used, for instance, to correct gender-based demographic biases in Recommender Systems \mbox{\cite{islam2021debiasing, islam2019mitigating}}.

    \item Another possibility would be to directly include EAUC as a fairness constraint in the dyadic regressor's optimizer during training; an interesting property of EAUC is that, unlike most AUC-based metrics, it is differentiable for any $\mathcal{D}$ with $|\mathcal{D}|\geq2$. Therefore, we could define a combined loss $\mathcal{L}_{fair}$ such that:

    \begin{align}
        \mathcal{L}_{fair}(B) &= \mathcal{L}_{MSE}(B) +\mathcal{L}_{EAUC}(B)\nonumber\\
        &= \frac{\sum\limits_{(u,i,r_{ui}) \in B}(r_{ui}-f(u,i))^2}{|B|}+EAUC(f,B) 
    \end{align}

    \noindent where $B \subseteq \mathcal{D}_{train}$ is a training batch. Similar one-step fair training paradigms have been used with more indirect parity and utility metrics by Yao and Huang \mbox{\cite{yao2017beyond}} to mitigate gender bias in movie recommendation, or through compositional adversarial learning by Bose and Hamilton \mbox{\cite{bose2019compositional}}, also in recommendation contexts.
    
    \item Yet another alternative would be to integrate EAUC as the fairness component of evaluation in multi-objective Auto-ML frameworks; recently, Freitas \mbox{\cite{freitas2024case}} proposed a conceptual framework based on the combination of Pareto and lexicographic approaches, where EAUC could be used as the main fairness criterion.
\end{enumerate}

In the following, we experimented with four simple, naive correction-based methods to demonstrate how lower EAUC correlates with less biased models. In the Netflix Prize dataset, we separate a portion of the Train set as the ``Correction'' set, which we use to compute different corrections $f_{corr}$ of the form  $\hat{r}_{ui_{\text{unbiased}}} = f_{corr}(\hat{r}_{ui_{\text{biased}}}, \bar{r}_u, \bar{r}_i)$:

\begin{itemize}
    \item A linear regression correction fitted to predict $\hat{r}_{ui_{\text{unbiased}}}$  as a function of $(\hat{r}_{ui_{\text{biased}}}, \bar{r}_u, \bar{r}_i)$.
    \item Two linear regression corrections with logit+clipping and sigmoid activations, respectively, that predict $\hat{r}_{ui_{\text{unbiased}}}$ as a function of $(\hat{r}_{ui_{\text{biased}}}, \bar{r}_u, \bar{r}_i)$, where examples have been uniformized with ML-RUS undersampling \mbox{\cite{charte2015addressing}} based on 10-bin discretized $\bar{r}_u$  and $\bar{r}_i$.
    \item A Random Forest correction with 100 decision trees and 10 layers of maximum depth, that predicts $\hat{r}_{ui_{\text{unbiased}}}$ as a function of $(\hat{r}_{ui_{\text{biased}}}, \bar{r}_u, \bar{r}_i)$.
\end{itemize}

\begin{figure}[h]
    \centering
    \includegraphics[width=.66\columnwidth]{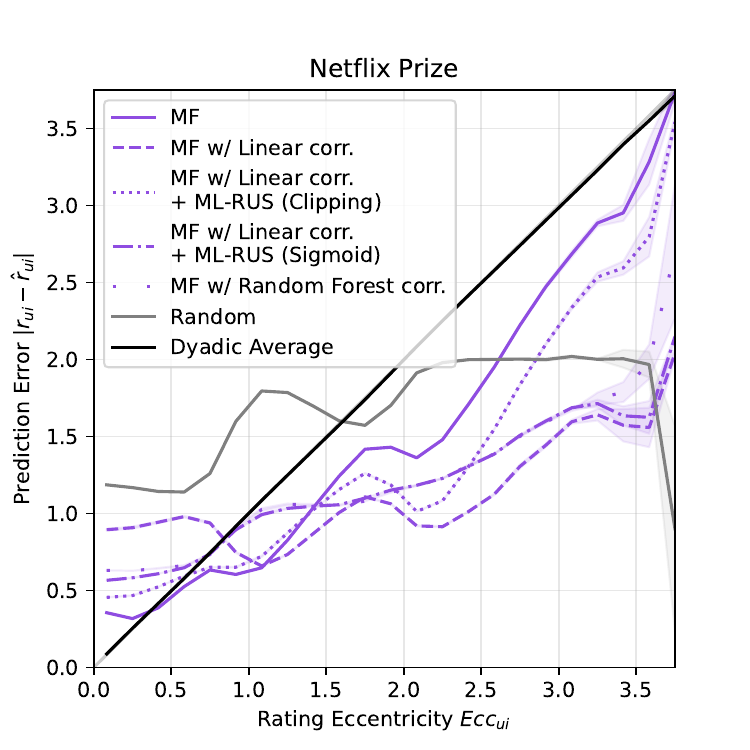}
    \caption{Relationship between the eccentricity of observed values and the prediction error (lower is better) for MF and naive correction de-basing techniques (avg. and std. dev. of 5 runs).}
    \label{fig:ecc_vs_error_graphs_corrs}
\end{figure}

Table \mbox{\ref{tab:correctionmetrics}} and Figure \mbox{\ref{fig:ecc_vs_error_graphs_corrs}} show the effects of the corrections in the global error (RMSE, MAE) and eccentricity bias (EAUC) and its curve analysis in an MF model and the Test set of Netflix Prize. Although they should not be interpreted as foolproof de-biasing corrections, their effects are evident. While they carry slightly higher RMSE than the original model on near-zero eccentricities, they maintain a much more restrained error as eccentricity increases; this, in turn, translates to a lower EAUC obtained by the correction, corresponding to lower eccentricity bias and less unfairness in contexts with such implications. Overall, this corroborates the adequacy of EAUC and its potential applicability to guide the construction of fair models.

\begin{table}[htbp]
\caption{Global error RMSE and MAE and eccentricity bias EAUC of MF and naive de-biasing corrections in Netflix Prize.}
\label{tab:correctionmetrics}
\resizebox{\columnwidth}{!}{%
\begin{threeparttable}
\setlength\extrarowheight{5pt}
\begin{tabular}{lrrr}
\toprule
 &
  \multicolumn{1}{l}{\textbf{RMSE}} &
  \multicolumn{1}{l}{\textbf{MAE}} &
  \multicolumn{1}{l}{\textbf{EAUC}} \\ \midrule
Random                 & 1.696 ± 0.000 & 1.386 ± 0.001 & 0.395 ± 0.004 \\
Dyad Average           & 0.964 ± 0.000 & 0.781 ± 0.000 & 0.435 ± 0.008 \\ \midrule
MF                     & \textbf{0.826 ± 0.001} & \textbf{0.635 ± 0.001} & 0.354 ± 0.006 \\ \midrule
w/ Linear corr.        & 1.186 ± 0.004 & 0.894 ± 0.003 & \textbf{0.253 ± 0.001} \\
\makecell[l]{w/ Linear corr. \\ + ML-RUS (Clipping)} &
  {\ul 0.879 ± 0.001} &
  {\ul 0.686 ± 0.001} &
  0.307 ± 0.001 \\
\makecell[l]{w/ Linear corr. \\ + ML-RUS (Sigmoid)} &
  0.977 ± 0.001 &
  0.784 ± 0.002 &
  {\ul 0.267 ± 0.001} \\
w/ Random Forest corr. & 1.008 ± 0.001 & 0.809 ± 0.001 & 0.276 ± 0.001 \\ \bottomrule
\end{tabular}

\begin{tablenotes}
\item \textit{For each metric best and second best results are bolded and underlined, respectively.}
\end{tablenotes}
\end{threeparttable}
}
\end{table}

\section{Conclusions and Future Directions}
\label{sec:conclusions}

In this study, we uncover a consisent bias in state-of-the-art models for dyadic regression, which we term \textit{eccentricity bias}, causing predictions to disproportionately gravitate toward the average observed values of users and items, and resulting in worse-than-random errors for more eccentric examples; we attribute this phenomenon to the non-uniformity in the local observed value distributions of entities. Furthermore, we prove that commonly used evaluation metrics like RMSE and MAE, which focus solely on global prediction errors, fail to capture eccentricity bias, raising fairness concerns about potential hazards in contexts with fairness requirements.

To address this limitation, we introduce Eccentricity-Area Under the Curve (EAUC) as a complementary metric capable of evaluating model performance across all levels of eccentricity in a given dyadic regression task. We demonstrate, in a comprehensive analysis with state-of-the-art models and real-world datasets on tasks with profound fairness implications (biomedicine, finances and product recommendation), that eccentricity bias is present in all studied contexts. We also prove that a lower EAUC correlates to performing acceptably well for all eccentricities rather than rewarding an overemphasis on predictions near the average observed values for users and/or items. Additionally, we show that an entity-wise adaptation of the Kolmogorov-Smirnov statistic can measure a dataset's susceptibility to induce eccentricity bias in models optimizing for global error minimization. Finally, we theorize on the potential usage of eccentricity bias not only as a means of evaluation, but also as a guideline to aid in the training and creation of fairer models, by experimenting with naive de-biasing corrections of trained unfair models.

As we look to the future, we believe the development of eccentricity bias-aware models and EAUC-based learning methodologies for dyadic regression should be of paramount importance. Consequently, as discussed earlier in this study, we outline this as the foremost further work to achieve fairness in contexts where it is critical for its societal impact.

\appendices
\section{Model Hyperparameters}
\label{apx:hyperparameters}
See Table \ref{tab:hyperparameters}.

\begin{table*}[h]
\setlength\extrarowheight{10pt}
\caption{Hyperparameter configurations per model and dataset.}
\label{tab:hyperparameters}
\scriptsize
\resizebox{\textwidth}{!}{
\begin{tabular}{llcccccc}
\hline
                        &                                                                              & \makecell{\textbf{Movielens} \\ \textbf{1M}} & \makecell{\textbf{Netflix} \\ \textbf{Prize}} & \textbf{GDSC1} & \textbf{CTRPv2} & \textbf{IMF DOTS}  & \textbf{Kiva} \\ \midrule
\textbf{MF}             & \makecell[l]{Embedding \\ Dimension}                & 512                     & 512                   & 512                    & 512                    & 8                     & 128             \\     
                        & \makecell[l]{Learning \\ Rate}                      & 0.001                  & 0.001                 & 0.001                  & 0.001                  & 0.001                 & 0.0005             \\
                        & \makecell[l]{$L_2$ Weight\\ Regularization}           & 0.00001                  & 0.00001               & 0.00001                & 0.00001                & 0.0001                & 0.0001               \\
                        &  \makecell[l]{Early Stop Patience}                                                         & 10                      & 10                    & 10                     & 10                     & 10                    & 10        \\           
                        & \makecell[l]{Early Stop Delta}                                                               & 0.0001                  & 0.0001                & 0.0001                 & 0.0001                 & 0.0001                & 0.0001     \\           \midrule
\textbf{Bayesian SVD++} & \makecell[l]{Embedding\\Dimension}                & 32                      & -                    & 10                    &    10                     &        10                &           10            \\
                        & \makecell[l]{MCMC\\Iterations}                     & 512                     & -                   & 512                    &  512                      &       512                 &         512              \\ \midrule
\textbf{GC-MC}          & Dropout                                                                      & 0.7                     & -                  & 0.7                    &   0.7                     & 0.7                   &      0.7                \\
                        & Hidden Layers                                                                & 500, 75                 & -               & 500, 75                &    500, 75                    & 500, 75               &        500, 75              \\
                        & \makecell[l]{Learning\\Rate}                       & 0.01                    & -                  & 0.01                   &         0.01               & 0.01                  &       0.01               \\
                        & \makecell[l]{Training\\Epochs}                     & 3500                    & -                  & 1000                 &    1000                    & 1000                   &    1000                  \\
                        & \makecell[l]{Mixture Model\\GCN Functions}         & 2                       & -                     & 2                      &     2                   & 2                     &   2                   \\ \midrule
\textbf{GLocal-K}       & \makecell[l]{Auto-Encoder\\Hiden Layers}          & 500, 500                & -              &      500, 500                  &        500, 500                & 500, 500              &    500, 500                  \\
                        & \makecell[l]{RBF\\Kernel Dimensions}               & 5, 5                    & -                 &   5, 5                     &    5, 5                    & 5, 5                  &   5, 5                   \\
                        & \makecell[l]{Global Kernel\\Dimension}             & 3x3                     & -                   &  3x3                      &   3x3                     & 3x3                   &      3x3                \\
                        & \makecell[l]{$\lambda_2$ $L_2$ Weight \\ Regularization}    & 70                      & -                    &     70                   &     70                   & 20                    &     70                 \\
                        & \makecell[l]{$\lambda_s$ $L_2$ Kernel \\ Regularization}    & 0.018                   & -                 &     0.018                    &      0.018                   & 0.006                 &     0.018                  \\
                        & \makecell[l]{L-BFGS-B Pre-training \\Iterations}  & 50                       & -                   &       5                 &            5            & 5                     &        5              \\
                        & \makecell[l]{L-BFGS-B Fine-tuning \\Iterations}  & 10                       & -                    &      5                  &             5           & 5                     &        5              \\ \bottomrule
\end{tabular}
}
\end{table*}
\section{Additional experiments on Recommender Systems}

This Appendix includes additional experiments (analysis of eccentricity-rating curves and evaluation of EAUC) in other popular Recommender Systems datasets: Movielens 100K and 10M \mbox{\cite{harper2015movielens}}, Douban Monti \mbox{\cite{monti2017geometric}}, and Tripadvisor \mbox{\cite{pereznunez2021}}. The three former and the latter involve user-movie and client-restaurant rating prediction, respectively

\begin{figure*}
    \centering
    \includegraphics[width=0.24\textwidth]{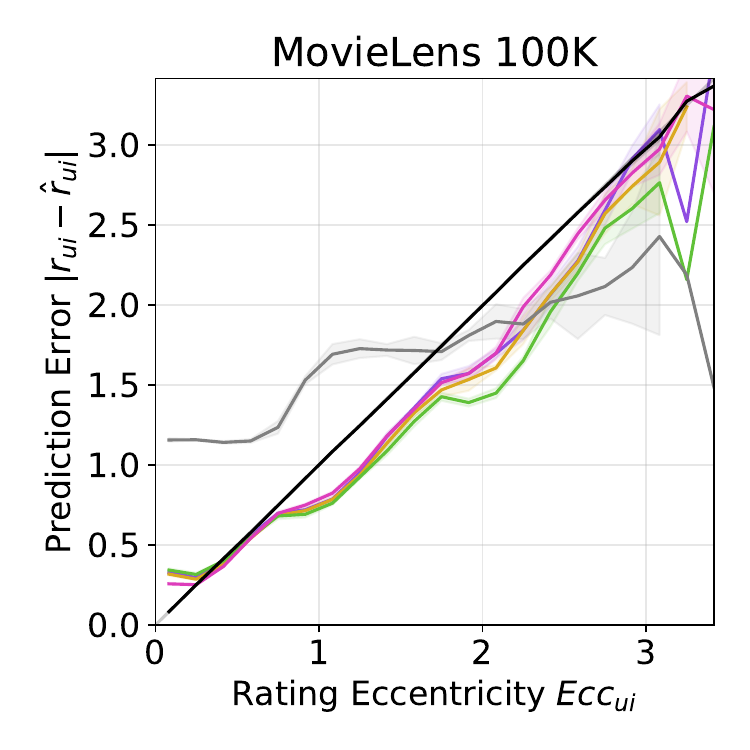}
    \includegraphics[width=0.24\textwidth]{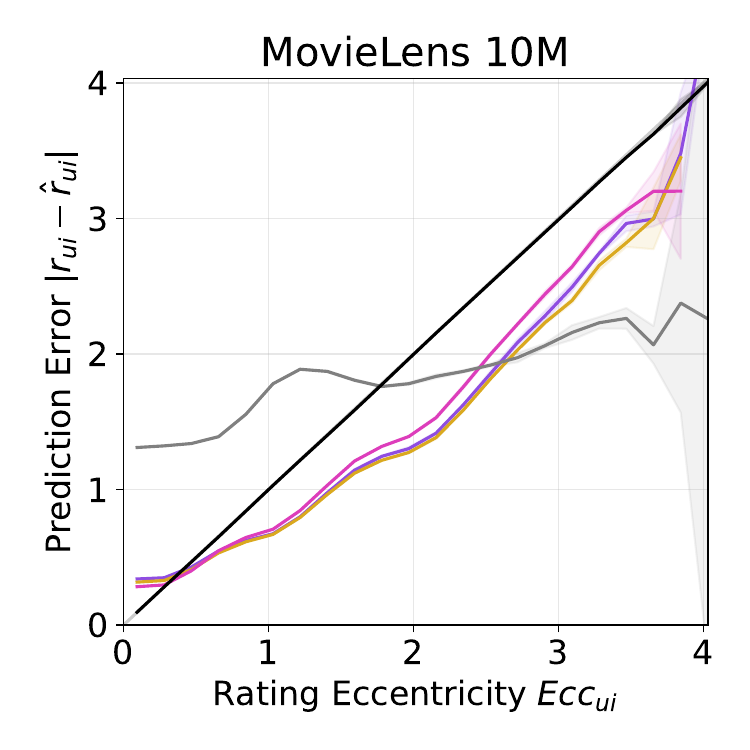}
    \includegraphics[width=0.24\textwidth]{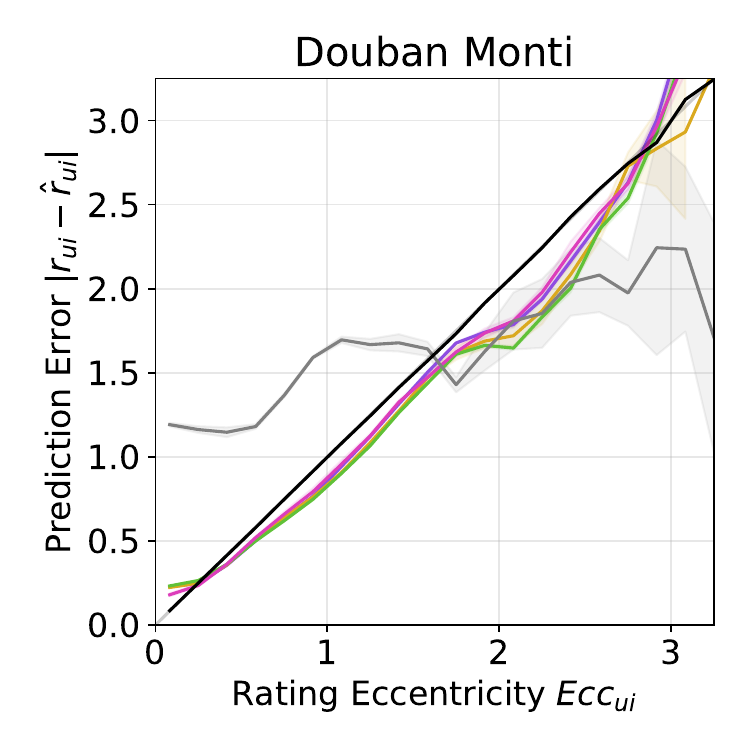}
    \includegraphics[width=0.24\textwidth]{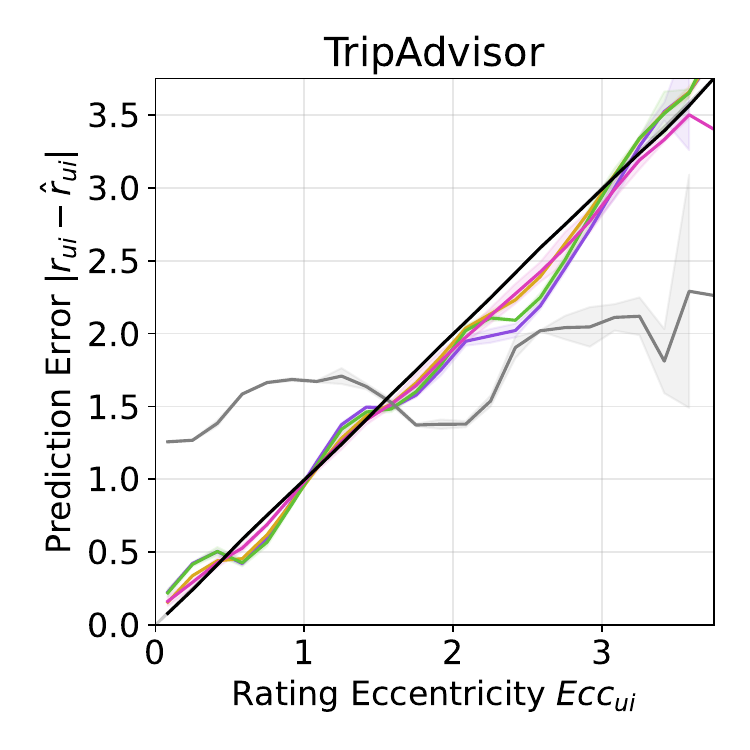}

    \caption{Relation between eccentricity observed values and prediction error (lower is better) for each model in additional Recommender Systems datasets (avg. and std. dev. of five runs). On Movielens 10M, GLOCAL-K could not be executed due to dataset size and model computational complexity.}
    \label{fig:curvesrecsys}
\end{figure*}

Table \mbox{\ref{tab:recsysdatasets}} contains basic dataset information and statistics, while Table \mbox{\ref{tab:resultsmetricsrecsys}} and Figure \mbox{\ref{fig:curvesrecsys}} showcase the eccentricity bias of ML models trained in each dataset. We hold the same conclusions as in results Section \mbox{\ref{sec:results}}: a lower RMSE or MAE does not necessarily correlate with a lower eccentricity bias, and RMSE or MAE cannot reliably detect eccentricity bias in model performance; contrarily, both EAUC and its curve representation can accurately describe and measure it in all models and datasets and provide a bias-aware component of performance in dyadic regression tasks.

\begin{table}[htbp]
\caption{Basic statistics of additional Recommendation dyadic regression datasets.}
\label{tab:recsysdatasets}
\resizebox{\linewidth}{!}%
{\begin{threeparttable}
\begin{tabular}{lrrrrrr}
\toprule
\textbf{Dataset} & \textbf{Examples}  & \textbf{Users} &\textbf{Items} & \textbf{Density} \\
\midrule
Movielens 100K& 100,000  & 944 & 1,683 & 6.30\% \\
Movielens 10M& 10,000,054  & 69,878 & 10,677 & 1.30\% \\
Douban Monti& 136,066  & 3,000 & 3,000 & 1.50\% \\
Tripadvisor& 491,170  & 25,388 & 12,384 & 0.20\% \\
\bottomrule
\end{tabular}
\begin{tablenotes}
\item \textit{Density indicates quantity of implicit feedback information available per entity (user or item) as $|\mathcal{D}| / (|\mathcal{U}| \times |\mathcal{I}|)$}
\end{tablenotes}
\end{threeparttable}
}

\end{table}

\begin{table*}[h]
\caption{RMSE, MAE, EAUC and $D_{KS_{\mathcal{D}}}$ per dataset and model in additional Recommendation dyadic regression datasets.}
\label{tab:resultsmetricsrecsys}
\begin{threeparttable}
\resizebox{\textwidth}{!}{%
\begin{tabular}{lrrrlrrr}
\toprule
 &
  \multicolumn{2}{l}{\textbf{Movielens 100K}} &
  \multicolumn{1}{l}{$D_{KS_{\mathcal{D}}}$: 0.415} &
   &
  \multicolumn{2}{l}{\textbf{Movielens 10M}} &
  \multicolumn{1}{l}{$D_{KS_{\mathcal{D}}}$: 0.437} \\ \midrule
 &
  \multicolumn{1}{l}{\textbf{RMSE}} &
  \multicolumn{1}{l}{\textbf{MAE}} &
  \multicolumn{1}{l}{\textbf{EAUC}} &
  \textbf{} &
  \multicolumn{1}{l}{\textbf{RMSE}} &
  \multicolumn{1}{l}{\textbf{MAE}} &
  \multicolumn{1}{l}{\textbf{EAUC}} \\ \midrule
Random       & 1.690 ± 0.009          & 1.381 ± 0.011          & 0.416 ± 0.015          &  & 1.841 ± 0.001       & 1.503 ± 0.000       & 0.409 ± 0.010       \\
Dyad Average & 0.978 ± 0.005          & 0.791 ± 0.005          & 0.401 ± 0.003          &  & 0.913 ± 0.001       & 0.719 ± 0.001       & 0.430 ± 0.007       \\ \cmidrule{1-4} \cmidrule{6-8} 
Bayesian SVD++ &
  {\ul 0.885 ± 0.005} &
  {\ul 0.693 ± 0.003} &
  {\ul 0.351 ± 0.006} &
   &
  \textbf{0.759 ± 0.001} &
  \textbf{0.580 ± 0.001} &
  \textbf{0.328 ± 0.011} \\
Glocal-K     & \textbf{0.882 ± 0.003} & \textbf{0.689 ± 0.004} & \textbf{0.343 ± 0.016} &  & -                   & -                   & -                   \\
GC-MC        & 0.891 ± 0.006          & 0.700 ± 0.005          & 0.371 ± 0.018          &  & {\ul 0.782 ± 0.001} & {\ul 0.594 ± 0.001} & 0.354 ± 0.013       \\
MF           & 0.907 ± 0.008          & 0.711 ± 0.007          & 0.365 ± 0.022          &  & {\ul 0.782 ± 0.001} & 0.597 ± 0.001       & {\ul 0.341 ± 0.013} \\ \bottomrule
\end{tabular}%
}
\resizebox{\textwidth}{!}{%
\begin{tabular}{lrrrlrrr}
\toprule
 &
  \multicolumn{2}{l}{\textbf{Douban Monti}} &
  \multicolumn{1}{l}{$D_{KS_{\mathcal{D}}}$: 0.466} &
   &
  \multicolumn{2}{l}{\textbf{Tripadvisor}} &
  \multicolumn{1}{l}{$D_{KS_{\mathcal{D}}}$: 0.539} \\ \midrule
 &
  \multicolumn{1}{l}{\textbf{RMSE}} &
  \multicolumn{1}{l}{\textbf{MAE}} &
  \multicolumn{1}{l}{\textbf{EAUC}} &
  \textbf{} &
  \multicolumn{1}{l}{\textbf{RMSE}} &
  \multicolumn{1}{l}{\textbf{MAE}} &
  \multicolumn{1}{l}{\textbf{EAUC}} \\ \midrule
Random       & 1.632 ± 0.011       & 1.333 ± 0.008          & 0.409 ± 0.019       &  & 1.826 ± 0.005       & 0.500 ± 0.005       & \textbf{0.425 ± 0.007} \\
Dyad Average & 0.771 ± 0.000       & 0.619 ± 0.000          & 0.416 ± 0.000       &  & 0.957 ± 0.004       & 0.737 ± 0.003       & 0.467 ± 0.020          \\ \cmidrule{1-4} \cmidrule{6-8} 
Bayesian SVD++ &
  \textbf{0.719 ± 0.009} &
  {\ul 0.564 ± 0.006} &
  \textbf{0.372 ± 0.004} &
   &
  \textbf{0.942 ± 0.005} &
  \textbf{0.725 ± 0.004} &
  0.453 ± 0.007 \\
Glocal-K     & {\ul 0.723 ± 0.001} & \textbf{0.562 ± 0.001} & {\ul 0.382 ± 0.002} &  & 0.964 ± 0.015       & 0.742 ± 0.015       & 0.457 ± 0.049          \\
GC-MC        & 0.732 ± 0.004       & 0.572 ± 0.006          & 0.399 ± 0.004       &  & {\ul 0.947 ± 0.012} & {\ul 0.733 ± 0.019} & 0.450 ± 0.029          \\
MF           & 0.735 ± 0.002       & 0.574 ± 0.003          & 0.399 ± 0.001       &  & 0.972 ± 0.008       & 0.753 ± 0.007       & {\ul 0.448 ± 0.020}    \\ \bottomrule
\end{tabular}%
}
\begin{tablenotes}
\item \textit{In each dataset, the first group of results  are naive baselines, and the second group are state-of-the-art models (avg. and std. dev. of 5 runs). Bold and underlined values represent the best and second best-performing model, respectively, of each dataset and metric. On Movielens 10M, GLOCAL-K could not be executed due to dataset size and model computational complexity.}
\end{tablenotes}
\end{threeparttable}
\end{table*}

\bibliographystyle{IEEEtran}
\bibliography{EAUC-IEEE}

\vfill

\end{document}